\documentclass[10pt,twocolumn,letterpaper]{article}

\usepackage{iccv}
\usepackage{times}
\usepackage{epsfig}
\usepackage{graphicx}
\usepackage{amsmath}
\usepackage{amssymb}
\usepackage{multirow}

\usepackage[pagebackref=true,breaklinks=true,letterpaper=true,colorlinks,bookmarks=false]{hyperref}

\iccvfinalcopy 

\def\iccvPaperID{2862} 

\ificcvfinal\fi

\begin{document}

\title{EgoCOL: Egocentric Camera pose estimation for Open-world 3D object Localization @Ego4D challenge 2023}

\author{Cristhian Forigua\\
Universidad de los Andes\\
{\tt\small cd.forigua@uniandes.edu.co}
\and
Maria Escobar\\
Universidad de los Andes\\
{\tt\small mc.escobar11@uniandes.edu.co}
\and
Jordi Pont-Tuset\\
Google Research Zürich \\
{\tt\small jponttuset@gmail.com}
\and
Kevis-Kokitsi Maninis\\
Google Research Zürich\\
{\tt\small  kevismaninis@gmail.com}
\and
Pablo Arbeláez\\
Universidad de los Andes\\
{\tt\small pa.arbelaez@uniandes.edu.co}
}

\maketitle
\ificcvfinal\thispagestyle{empty}\fi

\begin{abstract}
We present EgoCOL, an egocentric camera pose estimation method for open-world 3D object localization. Our method leverages sparse camera pose reconstructions in a two-fold manner, video and scan independently, to estimate the camera pose of egocentric frames in 3D renders with high recall and precision. We extensively evaluate our method on the Visual Query (VQ) 3D object localization Ego4D benchmark. EgoCOL can estimate 62\%  and 59\% more camera poses than the Ego4D baseline in the Ego4D Visual Queries 3D Localization challenge at CVPR 2023 in the val and test sets, respectively. Our code is publicly available at \url{https://github.com/BCV-Uniandes/EgoCOL}
\end{abstract}
\section{Introduction}
\label{sec:intro}
\begin{figure}
    \centering
    \includegraphics[width=\linewidth]{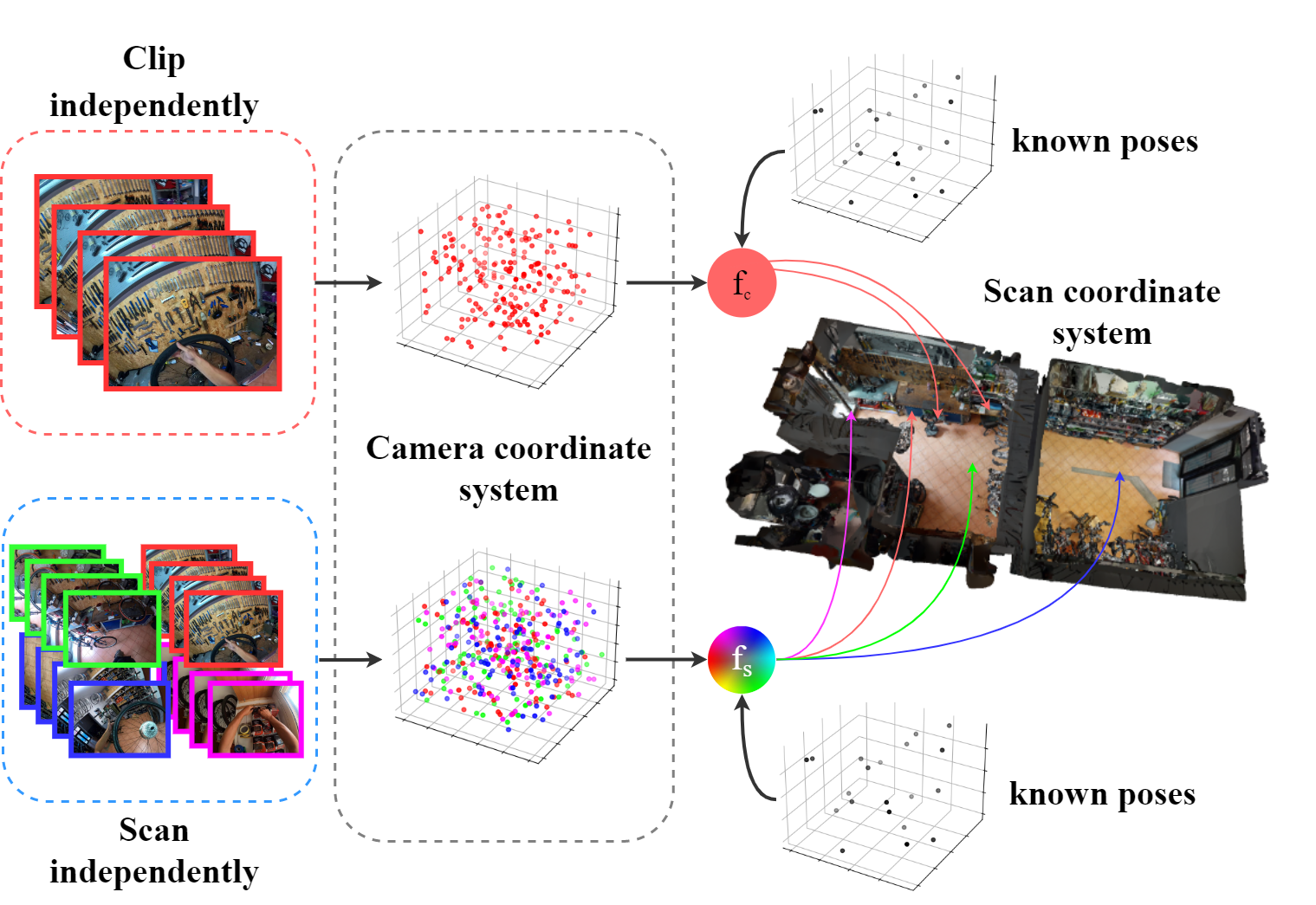}
    \caption{Overview figure of EgoCOL. \textbf{Camera pose estimation for egocentric videos.} We leverage both video and scan independently sparse relative camera pose reconstructions to compute transformations ($f_i,f_s$) that maps into the scan coordinate system. Our approach maps a higher number of frames into the 3D space by taking the most of a few known poses. Each color represents a different video.}
    \label{fig:overview}
\end{figure}

Most existing computer vision techniques aim at understanding the visual world from a third-person perspective \cite{yu2022coca,zhai2022scaling,liu2022swin,dai2021coatnet,liu2022swin,li2022grounded,dai2021dynamic,xu2021end,ran2022surface,wang2022multimodal,wang2022rbgnet,vu2022softgroup,liu2021group,duan2022revisiting, wei2022masked, yan2022multiview,girdhar2021anticipative,furnari2019would,wu2022memvit}. However, in both augmented reality and robotics applications, we need to understand data captured from a first-person or egocentric point-of-view. Through egocentric data, we see the world from the perspective of an agent interacting with its environment. Thus, egocentric perception requires a constant 3D awareness of the camera wearer's environment and must recognize objects and actions in human context \cite{grauman2022ego4d}. \\
\begin{figure}
  \centering    \includegraphics[width=\linewidth]{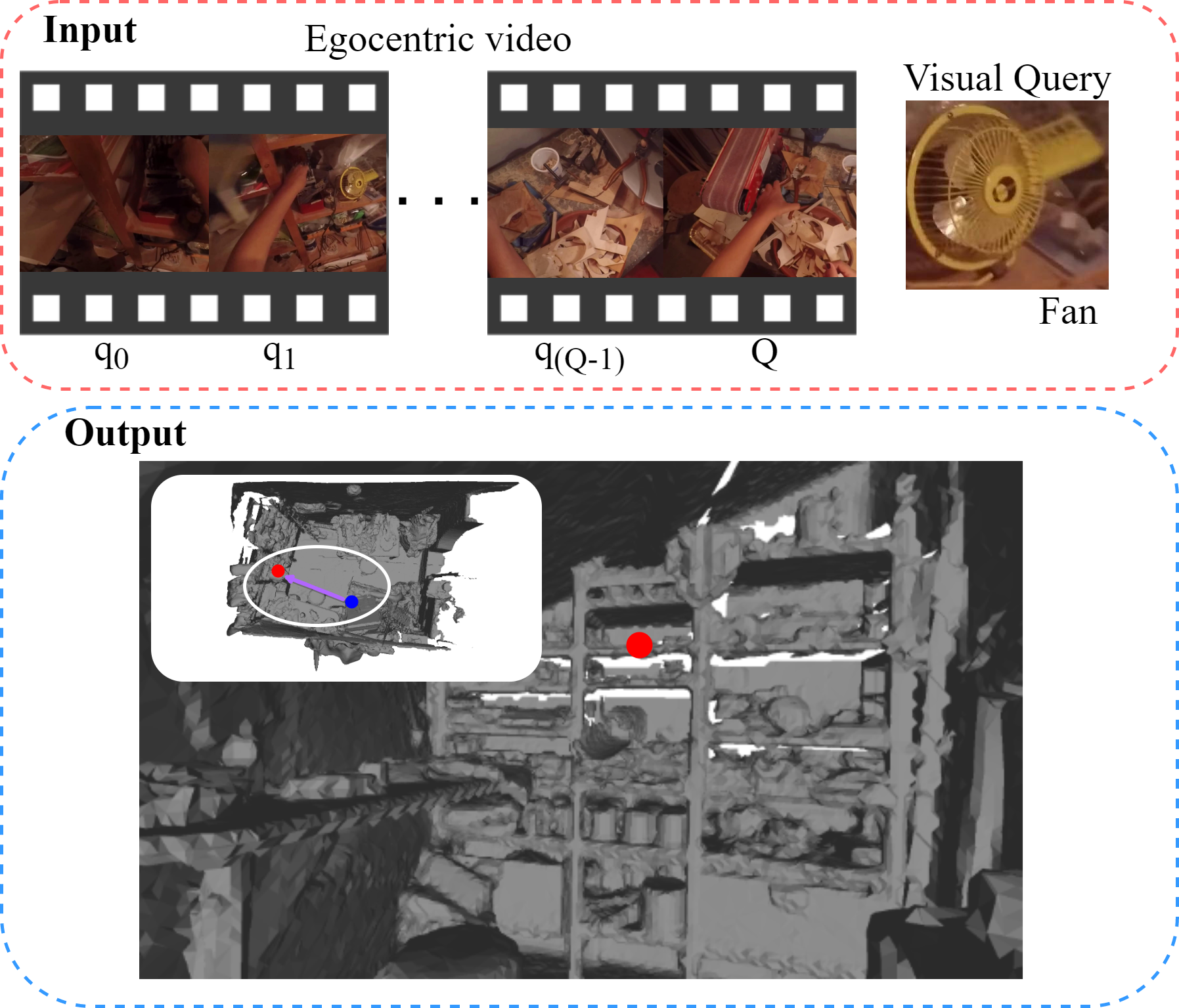}
  \caption{\textbf{Visual query 3D object localization.} Given an egocentric video, a query frame $Q$, and a visual query with 
 a target object, the goal is to locate a 3D displacement vector from the query frame $Q$ (\textit{blue point}) to the 3D location where the target object was last seen before $Q$ (\textit{red point}) in the egocentric video. EgoCOL processes an egocentric video and spatially locates the last appearance of the target object in the 3D scan.}
  \label{fig:VQ3D}
\end{figure}
Egocentric video understanding offers a variety of tasks centered around recognizing visual experience in the past, present, and future \cite{grauman2022ego4d,damen2016you,nagarajan2019grounded,kazakos2019epic,li2021ego,del2016summarization,ng2020you2me,yonetani2016recognizing,abu2018will,girdhar2021anticipative}. However, these methods rely on a 2D perception and fail to relate an egocentric experience to the 3D world. Since egocentric videos record daily life experiences such as \textit{who}, \textit{what}, \textit{when}, and \textit{where} of an individual, they are ideal for extending the 2D human perception into the 3D world for addressing tasks such as 3D episodic memory \cite{tulving1972episodic}. We define episodic memory by answering the question:\textit{ when and where I last saw object X?}.\\
In this report, we tackle the task of episodic memory with visual queries for 3D localization. As shown in Fig. \ref{fig:VQ3D}, given an egocentric video and a Visual Query (VQ), where the query is a static image of an object, we spatially locate in a 3D scene the last time the VQ was seen in the egocentric video. VQ 3D localization allows us to find an object of interest in a predefined 3D space through an egocentric video input. By working with visual queries, as opposed to categorical queries in which there is a fixed set of categories, VQ 3D localization is an approach for open-world 3D object localization, where no categories are needed for object detection.\\
We present EgoCOL, a new method for camera pose estimation from an egocentric perspective. EgoCOL leverages sparse reconstructions from video and scans to estimate the camera poses with high recall and precision. As shown in Fig. \ref{fig:overview}, we leverage sparse 3D relative camera pose reconstructions to compute transformation functions that allow us to map a large number of frames to the 3D scan coordinate system. In addition, we propose a to evaluate the VQ3D task a 3D detection problem. This approach helps mitigate the tendency of previous metrics to overestimate their performance.\\
\section{EgoCOL}
\subsection{Visual Query 3D Localization Task}\label{sec:vq3d}
Following Grauman et al. \cite{grauman2022ego4d}, we can formulate the task as follows. As shown in Fig. \ref{fig:VQ3D}, given an egocentric video $V$ of an interior room, a query object $o$ depicted in a static visual crop, and a query frame $q$, the objective is to identify the last 3D location of the object $o$ before the $q$.\\
We can separate Visual Query 3D localization into three sub-tasks: (\textit{1}) estimate the camera poses, (\textit{2}) locate the last occurrence $r_e$ of $o$ in $V$ and (\textit{3}) predict the location of $r_e$ in the 3D space. Then, the goal is to locate $r_e$ in the 3D scene coordinate systems to compute the displacement vector $(\Delta x, \Delta y, \Delta z)$ of the object with respect to the camera center at the query frame $q$.\\ Our method, which we refer to as EgoCOL, consists of multiple stages. We first estimate the relative camera poses of the video frames by using Structure from Motion~\cite{COLMAP} (SfM), and we register the cameras on the 3D world by using 7-DoF Procrustes \cite{sorkine2017least}. Then, we perform 3D object retrieval. Finally, we compute the 3D position through the displacement vector from the query frame.
\subsection{Camera Pose Estimation and Registration}\label{sec:poses}
Our approach consists of the following steps at a high level: First, we compute 3D SfM relative camera pose reconstructions by video and scan independently. For video configuration,  we compute an independent SfM reconstruction for each egocentric video. In addition, for the scan configuration, we merge all the videos that were capture on the same scan and we use them together to estimate the SfM reconstruction. Second, we estimate the camera poses by computing Procrustes mapping functions from the relative camera pose reconstructions to the scan coordinate system as shown in Fig. \ref{fig:overview}. These transformations are the best-fitting up-to-scale rigid transforms that aligns the SfM point with a set 3D world scan points. For Procrustes estimation, we compute a set of initial poses in the 3D world coordinate system through a keypoint matching strategy with a Perspective-n-Point (PnP) resolution approach. Lastly, we use the resulting 7 DoF transformation to map all the camera centers from the relative camera reconstructions to the 3D world of the scan. To merge the videos for the scan configuration, we take all the frames from the videos captured in the same scene and use those images to compute an SfM reconstruction for each scan. 
\subsubsection{Structure-from-Motion}
In order to increase the recall for the camera pose estimation, we first compute relative camera pose reconstructions. Since videos are recorded in indoor environments, we do not need to perform an extensive sampling to compute both 3D points and camera poses. Instead, we subsample each video to construct an initial sparse 3D reconstruction with COLMAP \cite{COLMAP}. This approach is more time efficient in terms of time and hardware requirements. Then, we register the remaining frames into the reconstructed model and perform a global bundle adjustment to refine the whole sparse model with the registered images. To compute the relative camera pose reconstructions under the scan configuration, we first merge the frames of all videos with the same camera intrinsic parameters. Then, we compute an initial sparse representation with a sub-set of the egocentric frames and register the remaining to get the final sparse reconstruction by each scan.
\subsubsection{Procrustes Mapping Transformation}
In order to register the  camera pose reconstructions into the 3D world scan coordinate system, we must compute a correspondence function between the two domains. Precisely, we use Procrustes mapping transform to find that correspondence.
Given two sets of corresponding points $\mathcal{P}=\{p_1, \ldots , p_n\}$ and $\mathcal{Q}=\{q_1, \ldots , q_n\}$ in $\mathbb{R}^{3}$, we compute a rigid transformation that aligns the two sets in the least squares sense \cite{sorkine2017least}. Precisely, we seek a rotation matrix $\mathcal{R}$, translation vector $\mathcal{T}$, and scale $s$ such that
\begin{equation*}
    (\mathcal{R},\mathcal{T}, s) = \operatorname*{argmin}_{\mathcal{R}\in SO(3), \mathcal{T}\in \mathbb{R}^{3}} \sum_{k=1}^{n}w_k||(\mathcal{R}(p_ks)+\mathcal{T})-q_k||^{2} 
\end{equation*}
where $w_i>0$ are weights for each point pair.\\
We first compute the weighted centroids of both point sets:
\begin{equation*}
     \hat{p}=\frac{\sum_{k=1}^{n}w_ip_i}{\sum_{k=1}^{n}w_i}, \hfill \hat{q}=\frac{\sum_{k=1}^{n}w_iq_i}{\sum_{k=1}^{n}w_i}
\end{equation*}
Then we compute the centered vectors:
\begin{equation*}
     x_k:=p_k-\hat{p} , \ \ \ \ y_k:=q_k-\hat{q} \ \ \ \ k=1,2, \ldots, n 
\end{equation*}
Now, we compute the $3 \times 3$ covariance matrix as
\begin{equation*}
     S = XWY^{T}
\end{equation*}
where $X$ and $Y$ are the $3 \times n$ matrices that have $x_i$ and $y_i$ as their columns, respectively, and $W$ is the a diagonal matrix with $(w1, w2, \dots , w_n)$. After, we compute the singular value decomposition (SVD) of $S$ as
\begin{equation*}
     SVD(S)=U\Sigma V^{T}
\end{equation*}
where $U$ is the matrix of the orthonormal eigenvectors of $SS^{T}$, $V$ the matrix that contains the orthonormal eigenvectors of $S^{T}S$ and $\Sigma$ is the diagnonal matrix containing the square roots of the eigenvalues of $S^{T}S$. Then, we can compute the rotation matrix $\mathcal{R}$ as
\begin{equation*}
\mathcal{R}=V
\begin{pmatrix}
1 &  & & \\
 &  \ddots & &\\
  &  &  1 & \\
   &  & &  det(VU^{T})\\
\end{pmatrix} U^{T}
\end{equation*}
and the translation vector $\mathcal{T}$ and scale $s$ as
\begin{equation*}
    \mathcal{T}=\hat{q}-\mathcal{R}p \ \ \ \ \ \ \ s= \frac{ \bar{y}}{max(\bar{x}, 1e^7)}
\end{equation*}
where $\bar{x}$ and $\bar{y}$ are the average distances to the centroids of each set of points respectively. In order to find the Procrustes transform from the relative camera pose reconstructions to the 3D world, we use a PnP approach to get an initial set of registered camera centers in the scan space as proposed in Grauman et al. \cite{grauman2022ego4d}.\\
Precisely, for camera pose estimation, we take $\mathcal{P}_v$ and $\mathcal{Q}_v$ as the common points computed by the PnP and sparse reconstruction approach in the video $v$, respectively. We then can use $\mathcal{R}$, $\mathcal{T}$ and $s$ to map all the registered points in the sparse reconstruction to the scan coordinate system by
\begin{equation*}
    f_c(P_i^{s}|\mathcal{R}_v,\mathcal{T}_v,s_v)= P_i^{s}(\mathcal{T}_v\mathcal{R}_cdiag(s_v))^{-1}
\end{equation*}
where $i$ refers to the i\textit{th} registered frame in the sparse reconstruction, $P_i^{s}$ is the camera pose at the i$th$ frame in the relative camera pose reconstruction space and $v$ is each independent video. For us to compute the transformation $f_v$, the set points $\mathcal{P}_v$ and $\mathcal{Q}_v$ must have at least three mutual (and non-co-linear) points, otherwise is not possible to compute the transformation. To solve this issue, we compute the transformation $f_s$ under the scan configuration by merging all the frames from the scene videos taken with the same camera device. This approach increases the number of mutual frames and fulfills the requirement in most cases. Our method leverages both $f_v$ and $f_s$ to estimate the camera poses with high recall. 
\subsubsection{Perspective-n-Point (PnP) Setup}
We follow the approach proposed by Grauman et al. \cite{grauman2022ego4d} to compute the initial PnP points. First, we estimate the intrinsic camera parameters. Second, we extract and match keypoints from the scans and the video frames. We use SuperPoint \cite{detone2018superpoint} for keypoint extraction and SuperGlue \cite{sarlin2020superglue} for matching. Then, we solve the set PnP setup and estimate camera poses from the matched pairs of 3D and 2D points with the estimated intrinsic camera parameters. Lastly, we add temporal constraints to compute non-localized camera poses and solve a new PnP to get the final pose estimation. Registering each frame independently using PnP is sub-optimal and insufficient to successfully compute camera poses with a high recall because of the sparsity of the correspondences in each frame. However, the PnP setup works to initiate the Procrustes transform estimation.
\subsection{3D Target Object Retrieval}
We evaluate our method using two different approaches for 3D target object retrieval. First, we follow the approach proposed by Grauman et al. \cite{grauman2022ego4d}. Second, we compute the 3D center of each scan and use it as the 3D predicted point for the target object. Finally, we compute the 3D displacement vector as proposed in \cite{grauman2022ego4d}.
\section{Experiments}
\begin{table*}[t!]
\centering
\resizebox{0.9\linewidth}{!}{%
\begin{tabular}{cl|c|cccc}
Rank & \multicolumn{1}{c|}{Method} & \textbf{Succ\%} $\uparrow$ & \textbf{Suc*\%} $\uparrow$ & \textbf{L2 $\downarrow$} & \textbf{angle} $\downarrow$ & \textbf{QwP\%} $\uparrow$ \\ \hline
5 & Ego4D-VQ3D-Baseline \cite{grauman2022ego4d} & 0.0795 & 0.4861 & 4.6377 & 1.3122 & 0.1629 \\
4 & thereisnospoon \cite{xu2022my} & 0.0909 & 0.5060 & 4.2339 & 1.2340 & 0.1629 \\
3 & Eivul Coming  \cite{mai2022estimating} & 0.2576 & 0.3874 & 8.9708 & 1.2045 & 0.6629 \\ 
2 & EgoCOL (ours) & 0.6288 & 0.8527 & 2.3718 & \textbf{0.5286} & 0.7462\\
\textbf{1} & VQ3D (EGO-LOC)& \textbf{0.8712} & \textbf{0.9614} & \textbf{1.8629} & 0.9205 & \textbf{0.9053} \\ \hline
\end{tabular}}
\caption{VQ 3D object localization results on the Ego4D benchmark in the test set. These results were obtained through the server available for the Ego4D Visual Queries 3D Localization challenge. The overall ranking is given by the Succ\% score. See the CVPR 2023 VQ 3D localization challenge leaderboard \href{https://eval.ai/web/challenges/challenge-page/1646/leaderboard/3947}{here}.}
\label{tab:sota}
\end{table*}
\subsection{Ego4D Benchmark}
We perform our experiments on the Ego4D benchmark for VQ 3D localization \cite{grauman2022ego4d}. The dataset consists of samples from 5 diverse scenarios with 3D scans. In total, 33 hours of video are distributed along 277 egocentric videos with 1032 visual queries. The dataset is divided into 604, 164, and 264 queries for train, validation, and test sets. We compute our results in the test set via an open server provided by the Ego4D VQ 3D localization challenge organizers. We report our results according to the metrics proposed by \cite{grauman2022ego4d}: root mean square error (RMSE), angular error (angle), success rate (Suc$c$\% and Suc$c^{*}$\% \footnote{Suc$c^{*}$ is the success metric computed only for queries with associated pose estimates \cite{grauman2022ego4d}.}), and the query ratio (QwP\%). To evaluate the capability of our method to recover the camera poses, we also present the percentages of frames and videos for which our method can compute the camera poses. Moreover, we assess the stringency of the challenge metric against the standard precision-recall methodology from the 3D object detection literature.
\subsection{Measuring the Metric}
We notice that the challenge metrics, especially the success rate, could overestimate the methods' performance. As shown in Eq. \ref{succ}, the success of a prediction depends on how close are the prediction $\hat{t_s}$ to the average centroid of the two 3D bounding boxes $c_m$ according to the threshold $6\times (||c_1-c_2||_2 + \delta)$. However, this threshold is too flexible and allows distances in the order of meters. Thus, high performance on the success rate does not implicate that the model predictions are close to the 3D bounding boxes annotations.
\begin{equation}
    succ = ||c_m - \hat{t_s}||_2<6\times (||c_1-c_2||_2 + \delta)
\label{succ}
\end{equation}
To gain further insights into the VQ3D task, we evaluate our results using standard precision-recall curves from the object detection literature \cite{vanrijsbergen1979information,abdou1979quantitative,martin2004learning,everingham2010pascal,lin2014microsoft,geiger2012we}. We threshold the overlap between the 3D bounding box of the annotation and the prediction, parameterize the precision-recall curve by the detector confidence and report the area under the curve or Average Precision (AP).\\
Table \ref{ab:ablationsval}  presents a comparison of the two performance metrics. While our best method demonstrates a twenty-four-fold improvement compared to the Ego4D baseline, achieving an overall success rate of 59\%, we observe that the AP is always zero regardless of the overlap threshold. Under closer inspection, \emph{we find a complete lack of overlap between the predictions and annotations}.
. This outcome suggests the use of standard precision-recall curves as complementary evaluation of the VQ3D task.
\subsection{State-of-the-art Comparison}
We evaluate EgoCOL against the Ego4D reference technique \cite{grauman2022ego4d} and the VQ 3D localization challenge participants at the Joint International 3rd Ego4D and 11th EPIC Workshop at CVPR 2023. Table \ref{tab:sota} shows the results for the Ego4D test set. EgoCOL came in second place at the CVPR 2023 Ego4D challenge.\\
In order to locate a target object in the 3D world, it is crucial to count the camera poses from the query frame and the frames in the response track so we can compute the 3D displacement vector. Regardless of the target object retrieval strategy, it is only possible to locate the detected object in the 3D world with the positions of the camera at an egocentric frame level. These results support that our main contribution, through which we compute camera poses with high recall and precision, is crucial to develop methods that need a 3D understanding of the world from a 2D egocentric perspective.\\
\begin{table*}[]
\centering
\resizebox{0.85\linewidth}{!}{%
\begin{tabular}{c|cc|cc}
\multirow{2}{*}{} & \multicolumn{2}{c|}{Test} & \multicolumn{2}{c}{Validation} \\
 & \% Videos $\uparrow$ & \% frames $\uparrow$ & \% Videos $\uparrow$ & \% frames $\uparrow$ \\ \hline
Eg4D \cite{grauman2022ego4d} & 63.77 & 14.74 & 63.64 & 6.26 \\ \hline
EgoCOL (video) & 34.09 & 26.83 & 47.83 & 37.21 \\
EgoCOL (scan) & \textbf{84.09} & 66.11 & 75.36 & 69.57 \\
EgoCOL (video+scan) & \textbf{84.09} & \textbf{66.67} & \textbf{84.06} & \textbf{75.97} \\ \hline
\end{tabular}}
    \caption{Camera pose estimation rate for the validation and test sets. We show the rate for videos and frames in each set. We compare against the state-of-the-art in the Ego4D Benchmark. Merging both video and scan configurations leads to the highest camera pose estimation rate for both sets. The rates were computed after filtering out the outliers.}
\label{tab:poses}
\end{table*}
\begin{table*}[t!]
\centering
\resizebox{\linewidth}{!}{%
\begin{tabular}{l|ccccc|c|}
\multicolumn{1}{c|}{Method} & \textbf{Succ\%} $\uparrow$ & \textbf{Suc*\%} $\uparrow$ & \textbf{L2} $\downarrow$ & \textbf{angle $\downarrow$} & \textbf{QwP\%} $\uparrow$ & \textbf{AP} $\uparrow$\\ \hline
Ego4D \cite{grauman2022ego4d} & 2.439 & 59.091 & 3.963 & 1.500 & 5.488 & 0.0 \\ \hline
EgoCOL (video) & 13.415 & 59.091 & 5.137 & 1.142 & 22.561 & - \\
EgoCOL (scan) & 6.707 & 55.556 & 6.189 & 0.964 & 12.195 & - \\
EgoCOL (video + scan) & 19.512 & 58.209 & 5.423 & 1.105 & 32.927 & - \\ \hline
EgoCOL (video + scan + filter) & 37.805 & 74.227 & 3.816 & 1.086 & 49.390 & - \\
\begin{tabular}[c]{@{}l@{}}EgoCOL (video + scan + 3D constrains)\end{tabular} & 52.439 & 80.165 & 3.527 & 0.967 & 63.415 & 0.0 \\ \hline
EgoCOL + 3D scan center & \textbf{59.146} & \textbf{93.388} & \textbf{2.306} & \textbf{0.579} & \textbf{63.415} & 0.0 \\ \hline
\end{tabular}}
\caption{Ablation experiments of EgoCOL on the validation set. EgoCOL uses the pretrained weights from \cite{grauman2022ego4d, xu2022negative} for the target object retrieval sub-task. The better results are in bold, and the second-highest results are underlined.}
\label{ab:ablationsval}
\end{table*}
\subsection{Camera Pose Estimation}
Table \ref{tab:poses} compares EgoCOL with the state-of-the-art method \cite{grauman2022ego4d} in terms of camera pose estimation. Our method outperforms \cite{grauman2022ego4d} by a high margin. We can estimate 60\% more frames in the validation set and 62\% more for the test set. We also compute camera poses for a higher fraction of the videos. We estimate camera poses for 22\% and 24\% more videos than Ego4D \cite{grauman2022ego4d} for the validation and test set, respectively. These results support that estimating the poses into the relative camera pose reconstruction space and mapping them to the 3D world works to construct a location correspondence between the camera and scan domains. EgoCOL solves the limitation of camera pose estimation by computing poses with a high recall. Thus, since camera pose estimation does not limit our method, we can locate more visual queries successfully, as shown in Tables  \ref{tab:sota} and \ref{ab:ablationsval}.\\
We perform an ablation study on the camera pose estimation in EgoCOL. Table \ref{tab:poses} shows the estimation rate by computing just the function $f_v$ for each video, $f_s$ for each set of videos captured in the same space with the same acquisition device and using both functions together. Our results demonstrate that using both functions leads to a higher estimation rate than using them separately. While the scan-independent approach shows a higher estimation rate across videos since it solves the mutual points limitation to compute the Procrustes mapping transform, the video-independent approach maps a higher number of poses in each video.

\subsection{Ablations}
Our ablation experiments thoroughly analyze EgoCOL's camera pose estimation configuration. We compare the performance of EgoCOL using video, scan, and both configurations to compute the camera poses of the egocentric frames. Moreover, we study the impact of post-processing the estimated camera poses by filtering out the outliers and applying 3D constraints: bringing the camera poses and final 3D predicted centroids outside the scan's 3D space to the closer vertex of the scan. Finally, we compare those approaches to taking the 3D center of each scan as the predicted 3D point. This last result motivated us to propose to evaluate the VQ3D task as a 3D detection problem. Table \ref{ab:ablationsval} shows the results of our ablations experiments.
\section{Conclusion and Limitations}
We present EgoCOL, an egocentric camera pose estimation method from an egocentric perspective for open-world 3D object localization. Our method significantly outperforms the Ego4D baseline, and we came second at the CVPR 2023 Ego4D challenge for VQ 3D Object localization. We find that leveraging sparse 3D relative camera pose reconstructions boosts VQ 3D object localization performance. A robust spatial correspondence between the 2D and 3D worlds is crucial to construct a sense of 3D perception from egocentric views effectively. Our method promotes the VQ 3D object localization task and could potentially boost downstream 3D tasks such as navigation and virtual reality applications.\\
Computing the correspondence between the 2D and egocentric 3D worlds is computationally expensive regarding hardware and execution time. In addition, it also depends on the length of the egocentric video since a higher number of frames would require more computational power and time. Despite EgoCOL implementing strategies to reduce these limitations (e.g., sub-sampling the video to estimate the relative camera pose reconstructions), the mapping between both domains is still time-consuming. Moreover, the camera poses estimation pipeline still suffers from outliers and requires post-processing to relocate the points and camera poses outside the scans. Finally, we observe a limitation in the stringency of the challenge metric, which we propose to overcome by considering standard average precision as a complementary evaluation

{\small
\bibliographystyle{ieee_fullname}
\bibliography{egbib}

\begin{thebibliography}{10}\itemsep=-1pt

\bibitem{abdou1979quantitative}
Ikram~E Abdou and William~K Pratt.
\newblock Quantitative design and evaluation of enhancement/thresholding edge
  detectors.
\newblock {\em Proceedings of the IEEE}, 67(5):753--763, 1979.

\bibitem{abu2018will}
Yazan Abu~Farha, Alexander Richard, and Juergen Gall.
\newblock When will you do what?-anticipating temporal occurrences of
  activities.
\newblock In {\em Proceedings of the IEEE conference on computer vision and
  pattern recognition}, pages 5343--5352, 2018.

\bibitem{dai2021dynamic}
Xiyang Dai, Yinpeng Chen, Bin Xiao, Dongdong Chen, Mengchen Liu, Lu Yuan, and
  Lei Zhang.
\newblock Dynamic head: Unifying object detection heads with attentions.
\newblock In {\em Proceedings of the IEEE/CVF conference on computer vision and
  pattern recognition}, pages 7373--7382, 2021.

\bibitem{dai2021coatnet}
Zihang Dai, Hanxiao Liu, Quoc~V Le, and Mingxing Tan.
\newblock Coatnet: Marrying convolution and attention for all data sizes.
\newblock {\em Advances in Neural Information Processing Systems},
  34:3965--3977, 2021.

\bibitem{damen2016you}
Dima Damen, Teesid Leelasawassuk, and Walterio Mayol-Cuevas.
\newblock You-do, i-learn: Egocentric unsupervised discovery of objects and
  their modes of interaction towards video-based guidance.
\newblock {\em Computer Vision and Image Understanding}, 149:98--112, 2016.

\bibitem{del2016summarization}
Ana~Garcia Del~Molino, Cheston Tan, Joo-Hwee Lim, and Ah-Hwee Tan.
\newblock Summarization of egocentric videos: A comprehensive survey.
\newblock {\em IEEE Transactions on Human-Machine Systems}, 47(1):65--76, 2016.

\bibitem{detone2018superpoint}
Daniel DeTone, Tomasz Malisiewicz, and Andrew Rabinovich.
\newblock Superpoint: Self-supervised interest point detection and description.
\newblock In {\em Proceedings of the IEEE conference on computer vision and
  pattern recognition workshops}, pages 224--236, 2018.

\bibitem{duan2022revisiting}
Haodong Duan, Yue Zhao, Kai Chen, Dahua Lin, and Bo Dai.
\newblock Revisiting skeleton-based action recognition.
\newblock In {\em Proceedings of the IEEE/CVF Conference on Computer Vision and
  Pattern Recognition}, pages 2969--2978, 2022.

\bibitem{everingham2010pascal}
Mark Everingham, Luc Van~Gool, Christopher~KI Williams, John Winn, and Andrew
  Zisserman.
\newblock The pascal visual object classes (voc) challenge.
\newblock {\em International journal of computer vision}, 88:303--338, 2010.

\bibitem{furnari2019would}
Antonino Furnari and Giovanni~Maria Farinella.
\newblock What would you expect? anticipating egocentric actions with
  rolling-unrolling lstms and modality attention.
\newblock In {\em Proceedings of the IEEE/CVF International Conference on
  Computer Vision}, pages 6252--6261, 2019.

\bibitem{geiger2012we}
Andreas Geiger, Philip Lenz, and Raquel Urtasun.
\newblock Are we ready for autonomous driving? the kitti vision benchmark
  suite.
\newblock In {\em 2012 IEEE conference on computer vision and pattern
  recognition}, pages 3354--3361. IEEE, 2012.

\bibitem{girdhar2021anticipative}
Rohit Girdhar and Kristen Grauman.
\newblock Anticipative video transformer.
\newblock In {\em Proceedings of the IEEE/CVF International Conference on
  Computer Vision}, pages 13505--13515, 2021.

\bibitem{grauman2022ego4d}
Kristen Grauman, Andrew Westbury, Eugene Byrne, Zachary Chavis, Antonino
  Furnari, Rohit Girdhar, Jackson Hamburger, Hao Jiang, Miao Liu, Xingyu Liu,
  et~al.
\newblock Ego4d: Around the world in 3,000 hours of egocentric video.
\newblock In {\em Proceedings of the IEEE/CVF Conference on Computer Vision and
  Pattern Recognition}, pages 18995--19012, 2022.

\bibitem{kazakos2019epic}
Evangelos Kazakos, Arsha Nagrani, Andrew Zisserman, and Dima Damen.
\newblock Epic-fusion: Audio-visual temporal binding for egocentric action
  recognition.
\newblock In {\em Proceedings of the IEEE/CVF International Conference on
  Computer Vision}, pages 5492--5501, 2019.

\bibitem{li2022grounded}
Liunian~Harold Li, Pengchuan Zhang, Haotian Zhang, Jianwei Yang, Chunyuan Li,
  Yiwu Zhong, Lijuan Wang, Lu Yuan, Lei Zhang, Jenq-Neng Hwang, et~al.
\newblock Grounded language-image pre-training.
\newblock In {\em Proceedings of the IEEE/CVF Conference on Computer Vision and
  Pattern Recognition}, pages 10965--10975, 2022.

\bibitem{li2021ego}
Yanghao Li, Tushar Nagarajan, Bo Xiong, and Kristen Grauman.
\newblock Ego-exo: Transferring visual representations from third-person to
  first-person videos.
\newblock In {\em Proceedings of the IEEE/CVF Conference on Computer Vision and
  Pattern Recognition}, pages 6943--6953, 2021.

\bibitem{lin2014microsoft}
Tsung-Yi Lin, Michael Maire, Serge Belongie, James Hays, Pietro Perona, Deva
  Ramanan, Piotr Doll{\'a}r, and C~Lawrence Zitnick.
\newblock Microsoft coco: Common objects in context.
\newblock In {\em Computer Vision--ECCV 2014: 13th European Conference, Zurich,
  Switzerland, September 6-12, 2014, Proceedings, Part V 13}, pages 740--755.
  Springer, 2014.

\bibitem{liu2022swin}
Ze Liu, Han Hu, Yutong Lin, Zhuliang Yao, Zhenda Xie, Yixuan Wei, Jia Ning, Yue
  Cao, Zheng Zhang, Li Dong, et~al.
\newblock Swin transformer v2: Scaling up capacity and resolution.
\newblock In {\em Proceedings of the IEEE/CVF Conference on Computer Vision and
  Pattern Recognition}, pages 12009--12019, 2022.

\bibitem{liu2021group}
Ze Liu, Zheng Zhang, Yue Cao, Han Hu, and Xin Tong.
\newblock Group-free 3d object detection via transformers.
\newblock In {\em Proceedings of the IEEE/CVF International Conference on
  Computer Vision}, pages 2949--2958, 2021.

\bibitem{mai2022estimating}
Jinjie Mai, Chen Zhao, Abdullah Hamdi, Silvio Giancola, and Bernard Ghanem.
\newblock Estimating more camera poses for ego-centric videos is essential for
  vq3d.
\newblock {\em arXiv preprint arXiv:2211.10284}, 2022.

\bibitem{martin2004learning}
David~R Martin, Charless~C Fowlkes, and Jitendra Malik.
\newblock Learning to detect natural image boundaries using local brightness,
  color, and texture cues.
\newblock {\em IEEE transactions on pattern analysis and machine intelligence},
  26(5):530--549, 2004.

\bibitem{nagarajan2019grounded}
Tushar Nagarajan, Christoph Feichtenhofer, and Kristen Grauman.
\newblock Grounded human-object interaction hotspots from video.
\newblock In {\em Proceedings of the IEEE/CVF International Conference on
  Computer Vision}, pages 8688--8697, 2019.

\bibitem{ng2020you2me}
Evonne Ng, Donglai Xiang, Hanbyul Joo, and Kristen Grauman.
\newblock You2me: Inferring body pose in egocentric video via first and second
  person interactions.
\newblock In {\em Proceedings of the IEEE/CVF Conference on Computer Vision and
  Pattern Recognition}, pages 9890--9900, 2020.

\bibitem{ran2022surface}
Haoxi Ran, Jun Liu, and Chengjie Wang.
\newblock Surface representation for point clouds.
\newblock In {\em Proceedings of the IEEE/CVF Conference on Computer Vision and
  Pattern Recognition}, pages 18942--18952, 2022.

\bibitem{sarlin2020superglue}
Paul-Edouard Sarlin, Daniel DeTone, Tomasz Malisiewicz, and Andrew Rabinovich.
\newblock Superglue: Learning feature matching with graph neural networks.
\newblock In {\em Proceedings of the IEEE/CVF conference on computer vision and
  pattern recognition}, pages 4938--4947, 2020.

\bibitem{COLMAP}
Johannes~L Schonberger and Jan-Michael Frahm.
\newblock Structure-from-motion revisited.
\newblock In {\em Proceedings of the IEEE conference on computer vision and
  pattern recognition}, pages 4104--4113, 2016.

\bibitem{sorkine2017least}
Olga Sorkine-Hornung and Michael Rabinovich.
\newblock Least-squares rigid motion using svd.
\newblock {\em Computing}, 1(1):1--5, 2017.

\bibitem{tulving1972episodic}
Endel Tulving.
\newblock Episodic and semantic memory.
\newblock 1972.

\bibitem{vanrijsbergen1979information}
C.~J. Van~Rijsbergen.
\newblock {\em Information Retrieval}.
\newblock Dept. of Computer Science, Univ. of Glasgow, 2nd edition, 1979.

\bibitem{vu2022softgroup}
Thang Vu, Kookhoi Kim, Tung~M Luu, Thanh Nguyen, and Chang~D Yoo.
\newblock Softgroup for 3d instance segmentation on point clouds.
\newblock In {\em Proceedings of the IEEE/CVF Conference on Computer Vision and
  Pattern Recognition}, pages 2708--2717, 2022.

\bibitem{wang2022rbgnet}
Haiyang Wang, Shaoshuai Shi, Ze Yang, Rongyao Fang, Qi Qian, Hongsheng Li,
  Bernt Schiele, and Liwei Wang.
\newblock Rbgnet: Ray-based grouping for 3d object detection.
\newblock In {\em Proceedings of the IEEE/CVF Conference on Computer Vision and
  Pattern Recognition}, pages 1110--1119, 2022.

\bibitem{wang2022multimodal}
Yikai Wang, Xinghao Chen, Lele Cao, Wenbing Huang, Fuchun Sun, and Yunhe Wang.
\newblock Multimodal token fusion for vision transformers.
\newblock In {\em Proceedings of the IEEE/CVF Conference on Computer Vision and
  Pattern Recognition}, pages 12186--12195, 2022.

\bibitem{wei2022masked}
Chen Wei, Haoqi Fan, Saining Xie, Chao-Yuan Wu, Alan Yuille, and Christoph
  Feichtenhofer.
\newblock Masked feature prediction for self-supervised visual pre-training.
\newblock In {\em Proceedings of the IEEE/CVF Conference on Computer Vision and
  Pattern Recognition}, pages 14668--14678, 2022.

\bibitem{wu2022memvit}
Chao-Yuan Wu, Yanghao Li, Karttikeya Mangalam, Haoqi Fan, Bo Xiong, Jitendra
  Malik, and Christoph Feichtenhofer.
\newblock Memvit: Memory-augmented multiscale vision transformer for efficient
  long-term video recognition.
\newblock In {\em Proceedings of the IEEE/CVF Conference on Computer Vision and
  Pattern Recognition}, pages 13587--13597, 2022.

\bibitem{xu2022negative}
Mengmeng Xu, Cheng-Yang Fu, Yanghao Li, Bernard Ghanem, Juan-Manuel Perez-Rua,
  and Tao Xiang.
\newblock Negative frames matter in egocentric visual query 2d localization.
\newblock {\em arXiv preprint arXiv:2208.01949}, 2022.

\bibitem{xu2022my}
Mengmeng Xu, Yanghao Li, Cheng-Yang Fu, Bernard Ghanem, Tao Xiang, and
  Juan-Manuel Perez-Rua.
\newblock Where is my wallet? modeling object proposal sets for egocentric
  visual query localization.
\newblock {\em arXiv preprint arXiv:2211.10528}, 2022.

\bibitem{xu2021end}
Mengde Xu, Zheng Zhang, Han Hu, Jianfeng Wang, Lijuan Wang, Fangyun Wei, Xiang
  Bai, and Zicheng Liu.
\newblock End-to-end semi-supervised object detection with soft teacher.
\newblock In {\em Proceedings of the IEEE/CVF International Conference on
  Computer Vision}, pages 3060--3069, 2021.

\bibitem{yan2022multiview}
Shen Yan, Xuehan Xiong, Anurag Arnab, Zhichao Lu, Mi Zhang, Chen Sun, and
  Cordelia Schmid.
\newblock Multiview transformers for video recognition.
\newblock In {\em Proceedings of the IEEE/CVF Conference on Computer Vision and
  Pattern Recognition}, pages 3333--3343, 2022.

\bibitem{yonetani2016recognizing}
Ryo Yonetani, Kris~M Kitani, and Yoichi Sato.
\newblock Recognizing micro-actions and reactions from paired egocentric
  videos.
\newblock In {\em Proceedings of the IEEE Conference on Computer Vision and
  Pattern Recognition}, pages 2629--2638, 2016.

\bibitem{yu2022coca}
Jiahui Yu, Zirui Wang, Vijay Vasudevan, Legg Yeung, Mojtaba Seyedhosseini, and
  Yonghui Wu.
\newblock Coca: Contrastive captioners are image-text foundation models.
\newblock {\em Transactions on Machine Learning Research}, 2022.

\bibitem{zhai2022scaling}
Xiaohua Zhai, Alexander Kolesnikov, Neil Houlsby, and Lucas Beyer.
\newblock Scaling vision transformers.
\newblock In {\em Proceedings of the IEEE/CVF Conference on Computer Vision and
  Pattern Recognition}, pages 12104--12113, 2022.

\end{thebibliography}
}

\end{document}